\documentclass[times, review, 10pt]{elsarticle}

\usepackage{amssymb}
\usepackage{amsmath}

\journal{Pattern Recognition}

\usepackage[ruled,linesnumbered]{algorithm2e}  
\DontPrintSemicolon            
\SetKwComment{tcc}{//}{}       

\usepackage{microtype}         
\usepackage{booktabs}        
\usepackage{multirow}        
\usepackage{makecell}        
\usepackage{longtable}       
\usepackage{adjustbox}
\usepackage{url}

\begin{document}

\begin{frontmatter}



\title{MCGM: Multi-stage Clustered Global Modeling for Long-range Interactions in Molecules}


\author[inst1]{Haodong Pan\fnref{equal}} 
\author[inst1]{Yusong Wang\fnref{equal}}
\author[inst1]{Nanning Zheng}
\author[inst1]{Caigui Jiang\corref{cor1}}

\fntext[equal]{These authors contributed equally to this work.}
\cortext[cor1]{Corresponding author. Email:cgjiang@xjtu.edu.cn}

\affiliation[inst1]{organization={State Key Laboratory of Human-Machine Hybrid Augmented Intelligence, National Engineering Research Center for Visual Information and Applications, and Institute of Artificial Intelligence and Robotics, Xi'an Jiaotong University},
            addressline={}, 
            city={Xi'an},
            postcode={710049}, 
            state={Shanxi},
            country={China}}

\begin{abstract}
Geometric graph neural networks (GNNs) excel at capturing molecular geometry, yet their locality-biased message passing hampers the modeling of long-range interactions. Current solutions have fundamental limitations: extending cutoff radii causes computational costs to scale cubically with distance; physics-inspired kernels (e.g., Coulomb, dispersion) are often system-specific and lack generality; Fourier-space methods require careful tuning of multiple parameters (e.g., mesh size, k-space cutoff) with added computational overhead. We introduce Multi-stage Clustered Global Modeling (MCGM), a lightweight, plug-and-play module that endows geometric GNNs with hierarchical global context through efficient clustering operations. MCGM builds a multi-resolution hierarchy of atomic clusters, distills global information via dynamic hierarchical clustering, and propagates this context back through learned transformations, ultimately reinforcing atomic features via residual connections. Seamlessly integrated into four diverse backbone architectures, MCGM reduces OE62 energy prediction error by an average of 26.2\%. On AQM, MCGM achieves state-of-the-art accuracy (17.0~meV for energy, 4.9~meV/\AA{} for forces) while using 20\% fewer parameters than Neural P\textsuperscript{3}M. Code will be made available upon acceptance.  
\end{abstract}


\end{frontmatter}



\section{Introduction}
The intersection of computational physics and pattern recognition has revolutionized molecular modeling, shifting from explicitly solving electronic structure equations to discovering hidden patterns in molecular interactions through deep learning, as systematically reviewed by Reiser et al. \cite{reiser2022graph}. Among these approaches, geometric graph neural networks (GNNs) \cite{batzner20223, ekstrom2023accelerating, schutt2021equivariant} have emerged as a central framework due to their natural compatibility with molecular topologies and spatial geometry. Unlike traditional Density Functional Theory (DFT) \cite{kohn1965self}, which requires iterative self-consistent field calculations, geometric GNNs enable direct end-to-end learning from data, providing rapid inference without expensive quantum mechanical computations. With $E(3)$-equivariant message passing and strong local geometric priors, GNNs can achieve near-DFT accuracy in energy and force predictions for molecular systems.

Current GNNs predominantly rely on local pattern extraction within fixed-radius neighborhoods, which becomes insufficient when long-range interactions are important for molecular properties. Although effective for small and compact molecules where all relevant interactions fall within typical cut-off radii (5-6~\AA{}, 0.5-0.6 nm), this approach struggles with extended molecular systems. For example, in molecules with more than 50 atoms-common in drug discovery and materials science-important interactions often extend beyond 10~\AA{}(1.0 nm)\cite{li2024longshortrange}. Increasing the cutoff radius may partially alleviate this issue, but causes computational costs to scale cubically \cite{gasteiger2020directional, kosmala2023ewald}. In contrast, ignoring long-range interactions introduces systematic errors \cite{cheng2025latent, staacke2021role}. Therefore, incorporating efficient and scalable long-range modeling mechanisms has become a key challenge in geometric molecular modeling.

To address the challenge of capturing long-range interactions beyond local neighborhoods, current approaches face distinct limitations: physics-based methods\cite{unke2019physnet, unke2021spookynet} incorporate explicit potentials but lack generality across chemical systems; Fourier-space methods\cite{kosmala2023ewald, cheng2025latent, wang2024neural} achieve global modeling but require significant computational overhead and complex parameter tuning; and existing hierarchical methods such as LSRM\cite{li2024longshortrange} employ fixed fragmentation rules that cannot adapt to diverse molecular topologies. In contrast, MCGM introduces adaptive hierarchical modeling through dynamic clustering, automatically discovering multi-scale patterns through efficient bounded operations-providing a practical and generalizable solution for long-range molecular interactions.

In this work, we propose \textbf{Multi-Stage Clustered Global Modeling (MCGM)}, a pioneering framework that fundamentally reimagines how molecular interactions are modeled hierarchically. Unlike all existing approaches that impose rigid structural assumptions—whether through physical equations, chemical rules, or spatial grids—MCGM achieves adaptive hierarchical decomposition through dynamic clustering in learned representation spaces. The key innovation lies in MCGM's ability to discover, rather than prescribe, molecular organization: at each training epoch, it dynamically reorganizes atoms into hierarchical clusters based on learned representations, discovering multi-scale structural patterns driven entirely by the learning objective without chemical priors. This dynamic hierarchy adapts not only to different molecular systems, but also evolves during training as the model learns richer representations, enabling the automatic discovery of task-relevant structural patterns that fixed hierarchies cannot capture. Remarkably, MCGM achieves this adaptive modeling through efficient hierarchical operations with a bounded computational cost per molecule, avoiding the computational overhead of Fourier-based methods like Neural P\textsuperscript{3}M\cite{wang2024neural} that require complex mesh constructions and parameter tuning. Moreover, its plug-and-play design allows seamless integration with any geometric GNN architecture without modifications, demonstrating that hierarchical global modeling can be both powerful and practical. This unique combination of adaptivity, efficiency, and modularity establishes MCGM as a new paradigm for hierarchical molecular interaction modeling.

To demonstrate MCGM's generality and plug-and-play nature, we integrate it into diverse GNN architectures spanning different design philosophies—from simple continuous filters (SchNet\cite{schutt2017schnet}) to angular-aware models (DimeNet++\cite{gasteiger2020fast}) and E(3)-equivariant architectures (PaiNN\cite{schutt2021equivariant}, GemNet-T\cite{gasteiger2021gemnet})—showing consistent improvements regardless of the underlying architecture. Our primary comparison is with Neural P\textsuperscript{3}M \cite{wang2024neural}, the current state-of-the-art for long-range molecular modeling, which uses the same set of backbones for a fair comparison. On OE62 \cite{stuke2020atomic}, MCGM achieves 26.2\% average improvement across these diverse architectures and outperforms Neural P\textsuperscript{3}M's best result (DimeNet++: 41.5 meV) by reaching 38.7 meV (6.2 $\times$ 10$^{-21}$J). On AQM \cite{medrano2024dataset}, MCGM with ViSNet establishes new state-of-the-art results (17.0 meV, 2.7 $\times$ 10$^{-21}$ J for energy; 4.9 meV/\AA{}, 7.8 $\times$ 10$^{-12}$ N for forces) while using 20\% fewer parameters than Neural P\textsuperscript{3}M. These results validate that MCGM not only seamlessly integrates with any GNN architecture but also consistently outperforms the best long-range modeling method currently.

\textbf{Our main contributions are as follows:}
\begin{itemize}
    \item \textbf{Adaptive hierarchical framework for molecular modeling.} We propose MCGM, which discovers multi-scale molecular organization through dynamic clustering in learned representation spaces. Unlike existing methods with fixed hierarchies, MCGM identifies structural patterns that evolve during training, enabling adaptive long-range interaction modeling without chemical priors.
    
    \item \textbf{Consistent performance improvements across architectures.} Extensive experiments demonstrate MCGM's effectiveness on both large-scale (OE62) and mid-scale (AQM) benchmarks, establishing new state-of-the-art results while maintaining parameter efficiency. The consistent gains across diverse backbone architectures validate MCGM's general applicability.
    
    \item \textbf{Universal plug-and-play design.} MCGM integrates seamlessly with both invariant (SchNet, DimeNet++) and equivariant (PaiNN, GemNet-T, ViSNet) architectures through a unified interface. The modular design requires minimal modifications to existing pipelines, facilitating adoption across different molecular modeling frameworks.
\end{itemize}

\section{Related Work}

\subsection{Geometric Graph Neural Networks}
Geometric GNNs model molecules as graphs with spatial symmetry constraints, categorized into SE(3)-invariant and SE(3)-equivariant architectures. Invariant models \cite{schutt2017schnet,gasteiger2020directional,gasteiger2021gemnet} operate on scalar quantities like distances and angles, providing rotational invariance but limited receptive fields. Equivariant models \cite{schutt2021equivariant,satorras2021n,musaelian2023learning,wang2024enhancing} maintain directional features for accurate force prediction. While effective for local interactions within cutoff radii (5-6 \AA), both approaches struggle with long-range effects in extended molecular systems where important interactions often exceed 10 \AA. To address these limitations, Graph kernel methods naturally provide hierarchical decompositions that complement GNNs' local operations \cite{xu2025graph}, motivating our adaptive clustering approach as an alternative to fixed kernel structures.

\subsection{Long-Range Interaction Modeling}
Building on these insights, various practical approaches have emerged to capture long-range interactions. Current methods can be categorized into two main strategies, each with inherent trade-offs.

\textbf{Physics-informed methods} incorporate explicit potentials: 4G-HDNNP \cite{ko2021fourth} uses QEq charge equilibration for electrostatics; Ewald-MP \cite{kosmala2023ewald} integrates Fourier-space summation; LSRM \cite{li2024longshortrange} employs chemical fragmentation with relay nodes. While achieving improvements, these methods require domain-specific knowledge and lack generality across chemical systems.

\textbf{Structural methods} enhance global awareness through architectural modifications: Neural-Atom \cite{li2023neural} and Equiformer \cite{liao2022equiformer} enable all-pair communications but with quadratic complexity; NLA-GNN \cite{wang2025nla} employs non-local attention to bypass local message-passing limitations; Graphormer \cite{ying2021transformers} uses full connectivity with learnable encodings; Neural P\textsuperscript{3}M \cite{wang2024neural} introduces FFT-compatible mesh nodes, achieving strong accuracy on benchmarks but requiring complex parameter tuning (mesh size, k-space cutoff). Recent hierarchical approaches \cite{lin2024deep,chen2024adaptive,ai2024a2gcn,guo2025collaborative} demonstrate the value of adaptive multi-scale modeling, yet rely on fixed decomposition rules that cannot adapt to diverse molecular topologies.

Unlike existing approaches that impose rigid assumptions—whether through physical equations, chemical rules, or spatial grids—MCGM achieves adaptive hierarchical decomposition through dynamic clustering in learned representation spaces. As shown in Table~\ref{tab:method_comparison}, MCGM uniquely combines adaptivity with plug-and-play modularity, automatically discovering task-relevant patterns while maintaining computational efficiency through bounded hierarchical operations.

\begin{table}[htbp]
    \caption{Comparison of long-range modeling approaches.}
    \label{tab:method_comparison}
    \centering
    \begin{tabular}{lccc}
    \toprule
    Method & Physics Prior & Adaptive & Plug-and-Play \\
    \midrule
    Extended Cutoff & No & No & Yes\\
    PhysNet/SpookyNet & Yes & No & No\\
    Ewald-MP & Yes & No & No \\
    Neural P\textsuperscript{3}M & No & No & Yes \\
    LSRM & No & No & No \\
    \textbf{MCGM (Ours)} & No & Yes & Yes \\
    \bottomrule
    \end{tabular}
\end{table}

\section{Methodology}

\subsection{Preliminaries}

\subsubsection{Problem Definition}
In molecular property prediction, a molecule is represented as $\mathcal{M} = \{Z, \mathbf{R}\}$, where $Z = \{z_1, \ldots, z_N\}$ denotes the atomic numbers and $\mathbf{R} = \{\mathbf{r}_1, \ldots, \mathbf{r}_N\}$ with $\mathbf{r}_i \in \mathbb{R}^3$ represents the 3D coordinates of $N$ atoms.

Throughout this paper, we adopt units standard in molecular modeling: distances in Ångströms (1 \AA= 0.1 nm = 10$^{-10}$ m), energies in millielectron volts (1 meV = 1.602 $\times$ 10$^{-22}$ J), and forces in meV/\AA (1 meV/\AA = 1.602 $\times$ 10$^{-12}$ N). These units, while not strictly SI, are universally accepted in computational chemistry and molecular physics literature.

The goal is to learn a function $\mathcal{F}_\theta$ that predicts molecular properties such as potential energy $E \in \mathbb{R}$ and atomic forces $\mathbf{F} \in \mathbb{R}^{3N}$, where forces are defined as the negative gradient of energy with respect to positions. The model is trained to minimize the discrepancy between predictions and reference values computed via DFT.

\subsubsection{Geometric Graph Neural Networks}
Geometric GNNs model a molecule as a graph $G = (V, E)$, where nodes $v_i \in V$ represent atoms and edges $e_{ij} \in E$ encode pairwise interactions within cutoff radius $d_c$. Each atom $i$ is characterized by its atomic number $z_i$ and position $\mathbf{r}_i \in \mathbb{R}^3$.

The node features are initialized as $\mathbf{h}_i^{(0)} = \text{Embed}(z_i)$ and updated through $L$ layers of message passing:
\begin{equation}
\mathbf{h}_i^{(\ell+1)} = \phi^{(\ell)}\left(\mathbf{h}_i^{(\ell)}, 
\bigoplus_{j \in \mathcal{N}(i)} \psi^{(\ell)}(\mathbf{h}_i^{(\ell)}, 
\mathbf{h}_j^{(\ell)}, e_{ij})\right)
\end{equation}
where $\mathcal{N}(i)$ denotes the neighbors of atom $i$ within cutoff $d_c$, $\bigoplus$ is a permutation-invariant aggregation (e.g., sum), and $\phi, \psi$ are learnable functions.

While effective for local interactions, this paradigm struggles with long-range effects beyond $d_c$. Although stacking more layers cloud theoretically expand the receptive field, this approach suffers from over-smoothing \cite{li2018deeper, oono2020graph} where node features become indistinguishable after excessive aggregation, and computational cost that scales linearly with the number of hops required. These limitations motivate our hierarchical extension that captures long-range interactions without deep stacking.

\subsection{Multi-stage Clustered Global Modeling}

\subsubsection{Overview}
\begin{figure*}[ht]
  \centering
  \includegraphics[width=\linewidth]{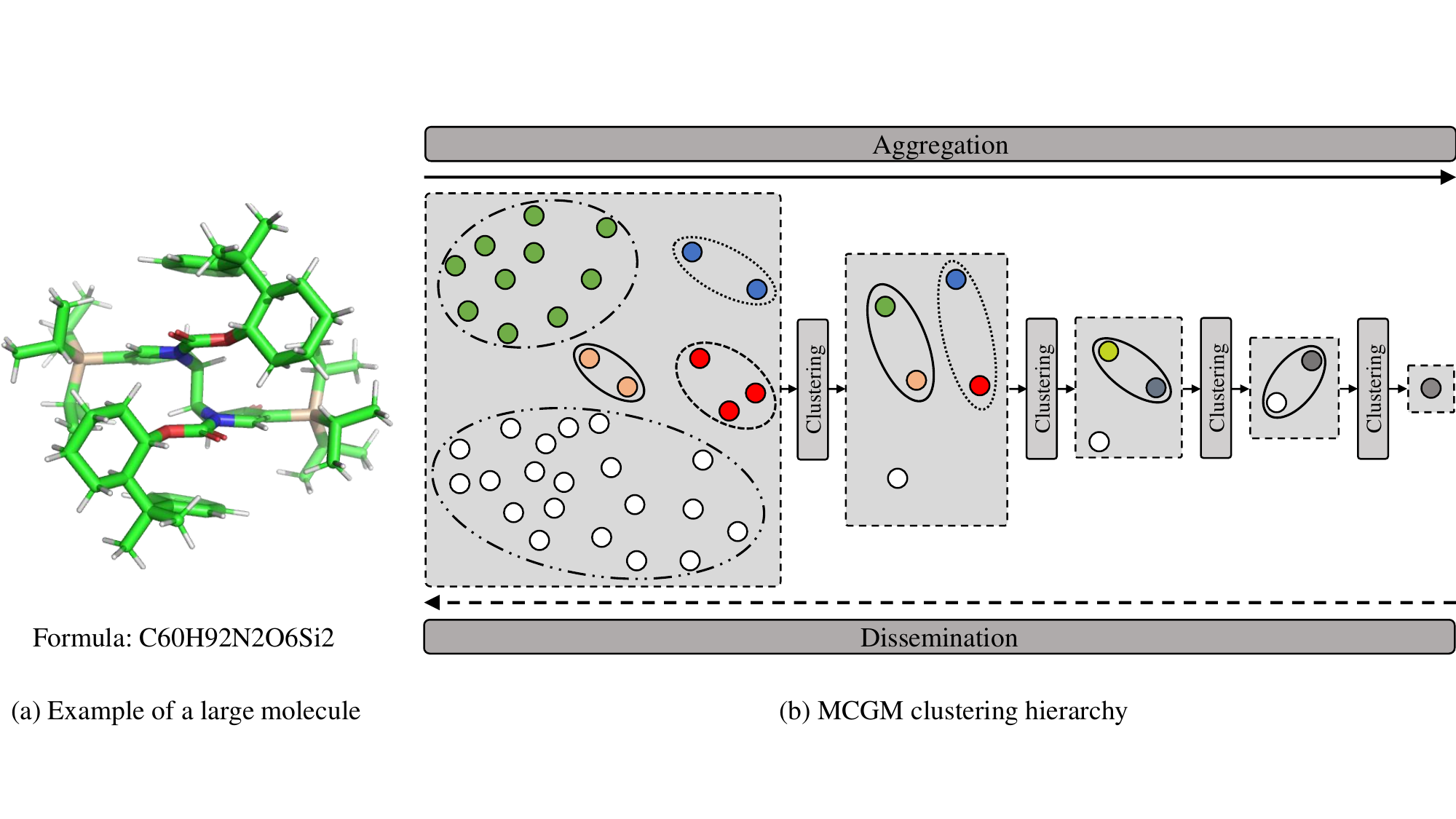}
  \caption{Overview of MCGM architecture. (a) Example molecule from the OE62 dataset illustrating the scale of structures handled by MCGM. (b) Multi-resolution clustering hierarchy. Dashed ellipses indicate clusters at each level, with nodes color-coded by cluster assignment. Grey boxes denote transitions between resolution levels.}
  \label{fig:mcgm_overview}
\end{figure*}

MCGM augments geometric GNNs with hierarchical global context through adaptive clustering in learned representation spaces. Given an atomic graph $G^{(0)}$ with $N$ atoms, MCGM constructs a hierarchy of progressively coarser graphs $\{G^{(1)}, ..., G^{(L)}\}$ where $|V^{(\ell+1)}| < |V^{(\ell)}|$. Information flows bidirectionally: local features are aggregated to coarser levels for global context extraction, then disseminated back to augment atomic representations. Unlike methods that rely on fixed hierarchies or physical priors, MCGM's clustering adapts dynamically during training based on learned embeddings, discovering task-relevant multi-scale patterns. The complete architecture is illustrated in Fig.~\ref{fig:mcgm_overview}, which demonstrates the hierarchical clustering process on a representative molecule.

\subsubsection{Hierarchical Graph Construction}
We construct a hierarchy of graphs $\{G^{(0)}, G^{(1)}, ..., G^{(L)}\}$ where $G^{(0)}$ is the original atomic graph with $N$ atoms, and each successive level represents progressively coarser clusters with $|V^{(\ell+1)}| < |V^{(\ell)}|$.

\textbf{Level 1 (Element-type clustering):} Atoms are grouped by element type, providing a strong chemical prior since atoms of different elements exhibit distinct interaction behaviors. This deterministic grouping creates $|V^{(1)}|$ clusters equal to the number of unique elements in the molecule.

\textbf{Level $\ell > 1$ (Adaptive clustering):} We apply K-means++ clustering \cite{Arthur2007kmeanspp} on the learned node embeddings from level $\ell-1$, with the number of clusters set to $|V^{(\ell)}| = \max(1, \lfloor |V^{(\ell-1)}|/r \rfloor)$ where $r$ is a reduction ratio (typically 2). Our hierarchical module then re-clusters nodes with a non-differentiable K-means++ algorithm, re-executed at each training epoch, providing an adaptive multi-level decomposition without extra learnable parameters or dense assignment matrices. This contrasts with DiffPool \cite{ying2018hierarchical}, which learns a differentiable soft assignment matrix through an additional pooling GNN at every layer, and with Cluster-GCN \cite{chiang2019cluster}, which relies on a single offline METIS partition purely to speed up mini-batch training rather than to learn a hierarchical representation.

\textbf{Edge construction:} Within each level $\ell$, we connect each node to its assigned cluster center, creating a star topology per cluster. This sparse connectivity pattern enables efficient information propagation between levels. The cluster centers serve as information hubs that aggregate local features and disseminate global context.

Through this hierarchical construction, the model captures interactions at multiple scales: local bonding at the atomic level, fragment-level patterns at intermediate levels, and global molecular context at the coarsest level.

\subsubsection{Information Flow Architecture}
We enable bidirectional information flow between hierarchical levels through two complementary operations: Aggregation (fine-to-coarse) and Dissemination (coarse-to-fine). This bidirectional exchange allows global patterns to inform local representations and vice versa.

\textbf{Aggregation} computes cluster features by pooling information from member nodes. For a cluster $C$ at level $\ell$ with members $\{i\}$ from level $\ell-1$, we compute:
\begin{equation} 
\mathbf{h}_C^{(\ell)} = W_{\text{agg}}^{(\ell)} \cdot \text{AvgPool}_{i \in C}
\left( [\mathbf{h}_i^{(\ell-1)} \| \phi(d_{iC})] \right)
\end{equation}
where $\|$ denotes concatenation, $d_{iC} = \|\mathbf{r}_i - \mathbf{r}_C\|$ is the Euclidean distance with cluster center position $\mathbf{r}_C = \frac{1}{|C|}\sum_{i \in C}\mathbf{r}_i$, and $\phi(\cdot)$ encodes distances using radial basis functions.

\textbf{Dissemination} propagates cluster information back to member nodes:
\begin{equation} 
\tilde{\mathbf{h}}_i^{(\ell-1)} = W_{\text{dis}}^{(\ell)} \cdot 
[\mathbf{h}_C^{(\ell)} \| \phi(d_{iC})]
\end{equation}

For the final dissemination from level 1 to the atomic level, we employ a residual connection to preserve local atomic features while incorporating global context:
\begin{equation}
\mathbf{h}_i^{(0)\text{final}} = \mathbf{h}_i^{(0)} + 
\tilde{\mathbf{h}}_i^{(0)}
\label{eq:dis_residual}
\end{equation}
This architecture enables efficient propagation of information across all scales, as each atom connects only to its assigned cluster center through a sparse star topology.

\subsubsection{Integration with GNN Backbones}
MCGM functions as a plug-and-play module that can augment any geometric GNN backbone. For invariant architectures (SchNet, DimeNet++), MCGM maintains strict invariance through scalar operations. For equivariant architectures (PaiNN, GemNet-T, ViSNet), MCGM enhances the scalar feature pathway while preserving the backbone's ability to process 
directional information. Force predictions are obtained through automatic differentiation of the energy with respect to atomic positions, maintaining equivariance in the output. The modular design requires minimal code modification to add hierarchical 
context to existing message passing.

\subsubsection{Energy and Force Prediction}
We employ hierarchical energy decomposition where atomic nodes and cluster centers independently contribute to the total energy. Two separate MLPs decode energy contributions: one processes atomic features to capture local interactions ($E_i$), another processes cluster features for collective effects ($E_C$). The total energy is:
\begin{equation}
E = \sum_{i \in \text{atoms}} E_i + \sum_{C \in \text{final clusters}} E_C
\end{equation}

This decomposition enhances interpretability by explicitly separating short-range (atomic) and long-range (cluster) contributions. Forces are obtained via automatic differentiation:
\begin{equation}
\mathbf{F}_i = -\frac{\partial E}{\partial \mathbf{r}_i}
\end{equation}
ensuring energy conservation and maintaining equivariance for force predictions even when using invariant features.

\section{Experiment}
\label{experiment}

\subsection{Experimental Setup}
\subsubsection{Datasets}
\textbf{OE62} \cite{stuke2020atomic} comprises approximately 62,000 organic molecules with 16 chemical elements and molecular sizes ranging from a few atoms to over 200. The dataset features molecules with spatial extents often exceeding 20 \AA{}, making long-range interactions crucial for accurate energy prediction. Ground-truth energies are 
computed using DFT with the PBE functional \cite{perdew1996generalized}. We adopt the same preprocessing and data split as in \cite{kosmala2023ewald}.

\textbf{AQM} \cite{medrano2024dataset} contains gas-phase conformations of drug-like molecules with up to 54 heavy atoms. We use the AQM-gas subset with 59,783 conformations from 1,653 molecules. Reference calculations employ DFTB3+MBD \cite{mortazavi2018structure}, incorporating many-body dispersion to accurately model long-range interactions. We use an 80/10/10 train/validation/test split with the same preprocessing pipeline as OE62.

\subsubsection{Baselines and Evaluation Metrics}
We evaluate MCGM against several categories of approaches for modeling long-range interactions:

\textbf{Architecture-agnostic improvements:}
(1) \textit{Embeddings}: Increasing hidden dimensions to enhance model capacity without architectural changes.
(2) \textit{Extended Cutoff}: Expanding the interaction radius, which increases computational cost cubically with distance.

\textbf{Specialized long-range methods:}
(3) \textit{SchNet-LR}: A variant introduced in \cite{kosmala2023ewald} that augments SchNet with a pairwise long-range block for interactions beyond the standard cutoff.
(4) \textit{Ewald MP} \cite{kosmala2023ewald}: Incorporates Ewald summation for long-range interactions through Fourier-space calculations.
(5) \textit{Neural P³M} \cite{wang2024neural}: Introduces learnable mesh nodes that propagate global information via FFT-compatible operations.
(6) \textit{Range variants} \cite{caruso2025extending}: Employ relaying attention nodes as intermediaries for global information exchange.

To demonstrate generality, we integrate MCGM with diverse GNN backbones: SchNet \cite{schutt2017schnet} (continuous filters), PaiNN \cite{schutt2021equivariant}~(equivariant), DimeNet++ \cite{gasteiger2020fast} (angular-aware), GemNet-T \cite{gasteiger2021gemnet} (higher-order invariants), and ViSNet \cite{wang2024enhancing} (lightweight equivariant). All models are evaluated using mean absolute error (MAE) on test sets. Baseline results are taken from their respective papers~\cite{kosmala2023ewald, wang2024neural, caruso2025extending}.

\subsubsection{Implementation Details}
MCGM employs K-means clustering with K-means++ initialization and a reduction ratio of 2 between hierarchical levels. Clustering is performed for 10 iterations with early stopping (tolerance 1e-4).

For OE62, we train models using L1 loss on energy predictions only. Optimization uses AdamW \cite{loshchilov2018decoupled} with learning rates of 5e-4 (SchNet), 1e-4 (PaiNN, DimeNet++), and 3e-4 (GemNet-T), with cosine annealing and warmup. Training uses early stopping (patience 50) and ReduceLROnPlateau scheduler (factor 0.8, patience 10).

For AQM, we jointly optimize energy and force predictions using weighted MSE loss with $\lambda_E = 0.01$ and $\lambda_F = 0.99$. Training uses AdamW with a learning rate of 1e-4, 10,000 warmup steps, and early stopping (patience 150). 

All models use a cutoff radius of 6.0 \AA{} for atomic interactions and 4.0 \AA{} for cluster interactions. Batch sizes range from 8-64 for OE62 and 32 for AQM. We run all experiments with three random seeds (0, 1, 2) and report mean ± standard deviation. Other hyperparameters follow the original papers.

\subsection{Main Results}
\subsubsection{Results on OE62}
To systematically assess MCGM's impact and generality, we integrate it into four widely used GNN architectures representing diverse geometric learning strategies: continuous-filter convolution (SchNet), equivariant message passing (PaiNN), explicit angular representations (DimeNet++), and higher-order invariants (GemNet-T).

Table~\ref{tab:oe62} presents the results. MCGM consistently improves all backbones, achieving an average relative improvement of 26.2\% on the OE62 test set. The improvement pattern is revealing: simpler architectures benefit most from MCGM's hierarchical modeling, with SchNet showing a 50.0\% improvement (131.3 to 65.6$\pm$0.8 meV), while sophisticated models like GemNet-T exhibit smaller but significant gains of 11.9\% (to 46.8$\pm$0.4 meV). This suggests MCGM effectively complements existing architectural innovations. DimeNet++-MCGM achieves the best absolute performance at 38.7$\pm$0.5 meV, 
surpassing the previous state-of-the-art Neural P\textsuperscript{3}M variant (41.5 meV).

Compared to other long-range approaches, MCGM demonstrates consistent advantages. The extended cutoff approach, while conceptually simple, shows limited effectiveness (e.g., SchNet-Cutoff: 254.8 meV vs baseline 131.3 meV), indicating that increased computational cost does not guarantee improved accuracy. More sophisticated methods like Ewald MP (81.1 meV) and SchNet-LR (89.2 meV) achieve meaningful improvements, yet MCGM (65.6 meV) surpasses them all. Notably, although Neural P\textsuperscript{3}M slightly edges out MCGM on PaiNN (52.9 vs 53.9$\pm$0.6 meV), MCGM achieves competitive accuracy with 24\% faster inference (1.64 vs 2.17 ms), offering a superior accuracy-efficiency trade-off.

Computationally, MCGM maintains efficiency across architectures. Inference overhead remains modest: 0.47 ms for SchNet (vs 0.13 baseline), 2.12 ms for DimeNet++ (vs 1.99), and 3.23 ms for GemNet-T (vs 3.07). These runtimes consistently match or outperform other long-range methods while achieving superior accuracy. The low standard deviations ($\le$0.8 meV) across three random seeds demonstrate the stability of MCGM's improvements.

\begin{table}[htbp]
  \centering
  \caption{Energy prediction MAE and computational cost on the OE62 dataset. Best-performing values are highlighted in \textbf{bold}. Energy units: 1 meV = 1.602 $\times$ 10$^{-22}$ J. Runtime is reported in milliseconds per molecular structure. Rel. indicates the relative improvement in percentage, and Fwd\,+\,Bwd refers to the total cost of a complete forward and backward pass.}
  \label{tab:oe62}

  \adjustbox{width=0.9\textwidth,center}{%
  
      \setlength{\tabcolsep}{1.5pt}
      \renewcommand{\arraystretch}{0.7} 

      \begin{tabular}{l c c c c c c c c c}
        \toprule
        \multirow{3}{*}{Model} & \multirow{3}{*}{Variant} &
        \multicolumn{2}{c}{OE62-val} &
        \multicolumn{2}{c}{OE62-test} &
        \multicolumn{2}{c}{Forward} &
        \multicolumn{2}{c}{Fwd\,+\,Bwd} \\
        \cmidrule(lr){3-4}\cmidrule(lr){5-6}\cmidrule(lr){7-8}\cmidrule(lr){9-10}
        & &
        \multicolumn{1}{c}{\makecell{MAE\\meV\,$\downarrow$}} &
        \multicolumn{1}{c}{\makecell{Rel.\\\%\,$\uparrow$}}  &
        \multicolumn{1}{c}{\makecell{MAE\\meV\,$\downarrow$}} &
        \multicolumn{1}{c}{\makecell{Rel.\\\%\,$\uparrow$}}  &
        \multicolumn{1}{c}{\makecell{Runtime\\ms\,$\downarrow$}} &
        \multicolumn{1}{c}{\makecell{Rel.\\\%\,$\downarrow$}} &
        \multicolumn{1}{c}{\makecell{Runtime\\ms\,$\downarrow$}} &
        \multicolumn{1}{c}{\makecell{Rel.\\\%\,$\downarrow$}} \\
        \midrule
        \multirow{6}{*}{SchNet}
          & Baseline      & 133.5 & {--}  & 131.3 & {--}  & 0.13 & {--}  & 0.28 & {--}  \\
          & Embeddings    & 144.7 & -8.4  & 136.7 & -4.1  & 0.14 & 15.2  & 0.33 & 17.8  \\
          & Cutoff        & 257.4 & -92.8 & 254.8 & -94.1 & 0.14 & 13.6  & 0.31 & 11.6  \\
          & SchNet-LR     & 86.6  & 35.1  & 89.2  & 32.1  & 0.32 & 156.0 & 0.75 & 171.7 \\
          & Ewald         & 79.2  & 40.7  & 81.1  & 38.2  & 0.70 & 461.6 & 1.03 & 271.4 \\
          & Neural P\textsuperscript{3}M & 70.2  & 47.4  & 69.1  & 47.4  & 0.37 & 184.6 & 0.57 & 103.6 \\
          & MCGM & \bfseries 66.6{\scriptsize$\pm$0.3} & \bfseries 50.1 & \bfseries 65.6{\scriptsize$\pm$0.8} & \bfseries 50.0 & 0.47{\scriptsize$\pm$0.01} & 261.5 & 1.05{\scriptsize$\pm$0.01} & 275.0 \\
        \midrule
        \multirow{6}{*}{PaiNN}
          & Baseline      & 61.4  & {--}  & 63.3  & {--}  & 1.52 & {--}  & 3.16 & {--}  \\
          & Embeddings    & 63.5  & -3.4  & 63.1  & -0.2  & 1.54 & 1.4   & 3.28 & 3.8   \\
          & Cutoff        & 65.1  & -6.0  & 64.4  & -2.2  & 1.84 & 20.9  & 3.91 & 23.6  \\
          & SchNet-LR     & 58.3  & 5.1   & 58.2  & 7.7   & 1.84 & 20.7  & 4.21 & 33.1  \\
          & Ewald         & 57.9  & 5.7   & 59.7  & 5.7   & 2.29 & 50.5  & 4.57 & 44.4  \\
          & Neural P\textsuperscript{3}M & \bfseries 54.1  & \bfseries 11.9  & \bfseries 52.9  & \bfseries 16.4 & 2.17 & 42.8  & 4.19 & 32.6  \\
          & MCGM & 56.3{\scriptsize$\pm$0.5}& 8.3 & 53.9{\scriptsize$\pm$0.6} & 14.8 & 1.64{\scriptsize$\pm$0.01} & 7.9 & 6.95{\scriptsize$\pm$0.19} & 119.9 \\
        \midrule
        \multirow{6}{*}{DimeNet++}
          & Baseline      & 51.2 & {--} & 53.8 & {--} & 1.99 & {--} & 4.26 & {--} \\
          & Embeddings    & 50.4 & 1.6  & 53.4 & 0.7  & 2.25 & 12.9 & 4.93 & 15.8 \\
          & Cutoff        & 48.3 & 5.7  & 48.1 & 10.6 & 2.68 & 34.7 & 6.10 & 43.4 \\
          & SchNet-LR     & 51.4 & -0.5 & 54.4 & -1.1 & 2.37 & 19.0 & 4.73 & 11.2 \\
          & Ewald         & 46.5 & 9.2  & 48.1 & 10.6 & 2.70 & 35.5 & 5.93 & 39.5 \\
          & Neural P\textsuperscript{3}M & 40.9 & 20.1 & 41.5 & 22.9 & 3.11 & 56.3 & 5.62 & 31.9 \\
          & MCGM & \bfseries 40.0{\scriptsize$\pm$0.3} & \bfseries 21.9 &
            \bfseries 38.7{\scriptsize$\pm$0.5} & \bfseries 28.1
            & 2.12{\scriptsize$\pm$0.02} & 6.5 & 8.01{\scriptsize$\pm$0.11} & 88.0 \\
        \midrule
        \multirow{6}{*}{GemNet‑T}
          & Baseline      & 51.5 & {--} & 53.1 & {--} & 3.07 & {--} & 6.96 & {--} \\
          & Embeddings    & 52.7 & -2.3 & 53.9 & -1.5 & 3.11 & 1.5  & 6.98 & 0.4  \\
          & Cutoff        & 47.8 & 7.2  & 47.7 & 10.2 & 4.02 & 31.2 & 8.88 & 27.7 \\
          & SchNet-LR     & 51.2 & 0.6  & 52.8 & 0.5  & 3.32 & 8.3  & 7.73 & 11.1 \\
          & Ewald         & 47.4 & 8.0  & 47.5 & 10.5 & 4.05 & 32.0 & 8.86 & 27.4 \\
          & Neural P\textsuperscript{3}M & \bfseries 47.2 & \bfseries 8.3  & 47.4 & 10.7 & 3.93 & 28.0 & 7.71 & 10.8 \\
          & MCGM & 48.7{\scriptsize$\pm$0.5} & 5.4 &
            \bfseries 46.8{\scriptsize$\pm$0.4} & \bfseries 11.9 & 3.23{\scriptsize$\pm$0.01} & 5.2 & 8.70{\scriptsize$\pm$0.35} & 25.0 \\
        \bottomrule
      \end{tabular}
    }
\end{table}

\subsubsection{Results on AQM}
For the AQM dataset, we evaluate MCGM integrated with ViSNet, a lightweight equivariant architecture optimized for force prediction—a critical requirement for this benchmark. Table~\ref{tab:aqm} presents the results.

ViSNet-MCGM achieves state-of-the-art performance on both tasks, with energy MAE of 17.0±0.3 meV and force MAE of 4.9±0.1 meV/\AA. This represents substantial improvements over existing long-range methods: 62.7\% better than SchNet-Ewald (45.6 meV) and 38.8\% better than SchNet-Range (27.8 meV) for energy prediction. Compared to the previous best method PaiNN-Range, ViSNet-MCGM reduces energy MAE by 12.8\% (from 19.5 to 17.0 meV) and force MAE by 36.4\% (from 7.7 to 4.9 meV/\AA).

The balanced performance across both energy and force predictions is particularly noteworthy. ViSNet-MCGM maintains excellent force accuracy (74.6\% improvement over SchNet-Ewald) while achieving the best energy performance, demonstrating MCGM's effectiveness in capturing both local gradients and global interactions. The larger relative improvements on AQM compared to OE62 (62.7\% vs 26.2\% average) suggest MCGM particularly excels when many-body dispersion interactions are dominant, as explicitly 
modeled in AQM's DFTB3+MBD reference calculations. These results validate MCGM's hierarchical approach for learning complex long-range effects in drug-like molecules with up to 54 heavy atoms.

\begin{table}[ht]
  \centering
  \caption{Energy MAE and Force MAE on the AQM dataset (best in \textbf{bold}). Energy units: 1 meV = 1.602 $\times$ 10$^{-22}$ J; Forces: 1 meV/\AA = 1.602 $\times$ 10$^{-12}$ N.}
  \label{tab:aqm}

    \begin{tabular}{l c c c c}
      \toprule
      \multirow{3}{*}{Method} & \multicolumn{2}{c}{Energy} 
      & \multicolumn{2}{c}{Forces}\\
      \cmidrule(lr){2-3}\cmidrule(lr){4-5}
      & \makecell{MAE\\meV\,$\downarrow$} & \makecell{Rel.\\\%\,$\uparrow$} &
          \makecell{MAE\\meV/\AA\,$\downarrow$} & \makecell{Rel.\\\%\,$\uparrow$}\\
      \midrule
      SchNet-Ewald & 45.6 & {--} & 19.3 & {--}\\
      SchNet-Range & 27.8 & 39.0 & 12.9 & 33.2\\
      PaiNN-Ewald  & 23.3 & 48.9 &  8.8 & 54.4\\
      PaiNN-Range  & 19.5 & 57.2 &  7.7 & 60.1\\
      ViSNet-MCGM  & \bfseries 17.0{\scriptsize$\pm$0.3} & \bfseries 62.7 &
                       \bfseries 4.9{\scriptsize$\pm$0.1} & \textbf{74.6}\\
      \bottomrule
    \end{tabular}
\end{table}

\subsection{Efficiency Analysis}
We evaluate the computational efficiency of ViSNet-MCGM against ViSNet-Neural P\textsuperscript{3}M's architecture on AQM. Table~\ref{tab:model_efficiency} reports model size and memory consumption.

ViSNet-MCGM uses 20\% fewer parameters than Neural P\textsuperscript{3}M (2.8 $\times$ 10 $^{6}$ vs 3.5 $\times$ 10 $^{6}$). While runtime memory reduction is modest (5.4\%) due to activation storage and framework overhead dominating total memory usage, fewer parameters provide important benefits: faster convergence during training, reduced overfitting risk, and simplified deployment.

The efficiency stems from architectural streamlining. Neural P\textsuperscript{3}M requires auxiliary mesh nodes with dedicated interaction layers and parameter-heavy continuous filters. In contrast, MCGM operates directly on adaptive cluster hierarchies using lightweight linear transformations, eliminating redundant components while improving predictive performance.

These results demonstrate that effective long-range modeling can be achieved through architectural simplicity rather than complexity, making MCGM particularly suitable for resource-conscious applications.

\begin{table}[ht]
  \centering
  \caption{Model size and memory usage on AQM. Measurements on RTX 4090 
  with batch size 1. Best values in \textbf{bold}.}
  \label{tab:model_efficiency}
  \adjustbox{width=0.98\textwidth}{
  \begin{tabular}{lccc}
    \toprule
    Model & Params & Train Mem. (MB) & Infer. Mem. (MB) \\
    \midrule
    ViSNet-Neural~P\textsuperscript{3}M & 3.5 $\times$ 10$^{6}$ & 362.8 & 361.2 \\
    ViSNet-MCGM & \textbf{2.8 $\times$ 10$^{6}$} & \textbf{343.1} & \textbf{341.6} \\
    \bottomrule
  \end{tabular}
}
\end{table}

\subsection{Ablation Studies}
We investigate the impact of clustering algorithms on MCGM's performance. While our hierarchical framework is designed to be flexible, the clustering quality affects the learned representations.

Table~\ref{tab:clustering_comparison} compares different clustering methods using SchNet-MCGM on OE62. K-means++ achieves the best performance (65.6 meV), with alternatives showing degraded accuracy: spectral (71.1 meV), random (72.9 meV), and random-balanced (74.9 meV). The 5.5-9.3 meV performance gaps demonstrate that while MCGM remains functional with various clustering schemes, K-means++ provides superior hierarchical decomposition.

This performance difference likely stems from K-means++ creating spatially coherent clusters that preserve local chemical environments while enabling effective cross-scale communication. Random clustering disrupts molecular structure, hindering the model's ability to learn meaningful hierarchical representations. These results validate our choice of K-means++ as an effective balance between computational efficiency and clustering quality.

\begin{table}[ht]
  \centering
  \caption{Impact of clustering algorithms on SchNet-MCGM energy prediction. 
Results on OE62 test set (MAE in meV). Energy: 1 \AA = 0.1 nm = 10$^{-10}$ m.}
  \label{tab:clustering_comparison}
  \begin{tabular}{lcccc}
    \toprule
    \multirow{2}{*}{Metric} & \multicolumn{4}{c}{Clustering Method} \\
    \cmidrule{2-5}
    & K-means++ & Spectral & Random Balanced & Random \\
    \midrule
    Energy & \textbf{65.6} & 71.1 & 74.9 & 72.9 \\
    \bottomrule
  \end{tabular}
\end{table}

\section{Conclusion}
\label{sec:conclusion}
We presented MCGM, a hierarchical framework that enhances long-range interaction modeling in molecular property prediction. By introducing adaptive multi-stage clustering with star-shaped Aggregation-Dissemination, MCGM enables efficient global context modeling through sparse hierarchical connections.

Our extensive evaluation demonstrates MCGM's effectiveness and generality. On OE62, MCGM improves four diverse GNN architectures by an average of 26.2\%, with DimeNet++-MCGM achieving 38.7 meV MAE—surpassing the previous state-of-the-art. On AQM, ViSNet-MCGM attains 17.0 meV for energy and 4.9 meV/\AA for forces, outperforming all baselines while using 20\% fewer parameters than comparable methods. The consistent improvements across different backbones (SchNet, PaiNN, DimeNet++, GemNet-T, ViSNet) confirm MCGM's plug-and-play applicability.

MCGM's design philosophy—achieving long-range modeling through architectural simplicity rather than auxiliary structures—offers a practical path for incorporating global interactions into existing molecular simulation pipelines. Beyond molecular modeling, the hierarchical clustering framework may benefit other domains requiring multi-scale spatial modeling, such as point cloud processing and 3D shape understanding. Future work could explore learnable clustering strategies, investigate community-aware pre-training approaches \cite{huang2025efficient} to enhance structural awareness, and validate MCGM's effectiveness across these diverse applications.

\appendix
\section{Algorithm Details}
\setcounter{algocf}{0}  
\renewcommand{\thealgocf}{A.\arabic{algocf}}
\subsection{K-means Clustering}
Algorithm~\ref{alg:A1} presents the hierarchical graph clustering procedure used in MCGM.

\begin{algorithm}[htbp]
\footnotesize 
\caption{Hierarchical Graph Clustering}
\label{alg:A1}
\KwIn{
  Node features $X\!\in\!\mathbb{R}^{N\times d}$; batch index $b$;\\
  reduction ratio $r=2$; tolerance $\varepsilon=10^{-4}$; max iterations $T=10$.}
\KwOut{Cluster assignments $c\!\in\!\{0,\dots,K-1\}^N$.}
\BlankLine
$B \gets \text{number of graphs in batch}$\;
\ForEach{graph $g$ in batch}{
  $n_g \gets \text{number of nodes in graph } g$\;
  $k_g \gets \lceil n_g/r \rceil$ \tcp{Clusters for this graph}
  Initialize centroids $C_g$ using K-means++ for graph $g$\;
}
\For{$t=1$ \KwTo $T$}{
  $C_{\text{old}} \gets C$\;
  \ForEach{node $i$}{
     Assign $i$ to nearest centroid in its graph\;
  }
  Update centroids as cluster means\;
  \If{any cluster is empty}{
     Reinitialize empty centroid randomly\;
  }
  \If{all clusters have single node}{\Return{$c$} \tcp{Early stop}}
  \If{$\|C - C_{\text{old}}\| \le \varepsilon$}{\Return{$c$}}
}
\Return{$c$}
\end{algorithm}

\subsection{Aggregation and Dissemination Units}
Algorithms~\ref{alg:A2} and \ref{alg:A3} detail the aggregation and dissemination procedures.

\begin{algorithm}[htbp]
\footnotesize
\caption{Aggregation Unit}
\label{alg:A2}
\KwIn{%
  Node features $F_f\!\in\!\mathbb{R}^{N\times d}$;
  positions $P_f\!\in\!\mathbb{R}^{N\times3}$;\\
  cluster assignments $c\!\in\!\{0,\dots,K-1\}^N$; total clusters $K$;\\
  cutoff radius $\gamma$; number of RBFs $N_R$.}
\KwOut{%
  Cluster features $F_c\!\in\!\mathbb{R}^{K\times d'}$;
  centroids $P_c\!\in\!\mathbb{R}^{K\times3}$.}
\BlankLine
\tcp{Compute cluster centroids}
\For{$k = 0$ \KwTo $K-1$}{
  $\mathcal{I}_k \gets \{i : c_i = k\}$\;
  \lIf{$\mathcal{I}_k = \varnothing$}{\textbf{continue}}
  $P_{c,k} \gets \frac{1}{|\mathcal{I}_k|}\sum_{i \in \mathcal{I}_k} P_{f,i}$\;
}
\tcp{Aggregate features to clusters}
\For{$k = 0$ \KwTo $K-1$}{
  $\mathcal{I}_k \gets \{i : c_i = k\}$\;
  \If{$\mathcal{I}_k \neq \varnothing$}{
    \ForEach{$i \in \mathcal{I}_k$}{
      $r_i \gets \lVert P_{f,i} - P_{c,k} \rVert_2$\;
      $e_i \gets \mathrm{RBF}(r_i;\gamma,N_R)$\;
      $z_i \gets [F_{f,i} \,\|\, e_i]$\;
    }
    $F_{c,k} \gets W_{\text{agg}} \cdot \frac{1}{|\mathcal{I}_k|}\sum_{i \in \mathcal{I}_k} z_i + b$\;
  }
}
\Return{$(F_c, P_c)$}
\end{algorithm}

\begin{algorithm}[htbp]
\footnotesize
\caption{Dissemination Unit}
\label{alg:A3}
\KwIn{%
  Cluster features $F_c\!\in\!\mathbb{R}^{K\times d}$;
  cluster positions $P_c\!\in\!\mathbb{R}^{K\times3}$;\\
  node positions $P_f\!\in\!\mathbb{R}^{N\times3}$;
  cluster assignments $c\!\in\!\{0,\dots,K-1\}^N$;\\
  cutoff radius $\gamma$; number of RBFs $N_R$.}
\KwOut{%
  Updated node features $F_f'\!\in\!\mathbb{R}^{N\times d'}$.}
\BlankLine
\tcp{Disseminate cluster information to nodes}
\For{$i = 0$ \KwTo $N-1$}{
  $k \gets c_i$ \tcp{Get cluster index for node $i$}
  $r_i \gets \lVert P_{f,i} - P_{c,k} \rVert_2$\;
  $e_i \gets \mathrm{RBF}(r_i;\gamma,N_R)$\;
  $z_i \gets [F_{c,k} \,\|\, e_i]$\;
  $F_{f,i}' \gets W_{\text{dis}} \cdot z_i + b$\;
}
\Return{$F_f'$}
\end{algorithm}

\section{Integration with GNN Backbones}
\setcounter{table}{0} 
MCGM seamlessly integrates with existing GNN architectures by introducing hierarchical clustering and cross-scale communication modules. The integration preserves the original architecture's properties while enabling long-range interactions.

\subsection{SchNet}
SchNet uses continuous-filter convolutions with radial basis functions. MCGM integrates after each SchNet layer, where atomic features $h^{(l)}$ and cluster features $h_C^{(l)}$ interact through aggregation-dissemination:
\begin{align}
h^{(l+1)} &= h^{(l)} + \Delta h_{\text{SchNet}}^{(l)} + W_{\text{dis}} \cdot [h_C^{(l)} \| e_{Ci}] \\
h_C^{(l+1)} &= h_C^{(l)} + W_{\text{agg}} \cdot \text{mean}_{i \in C}([h_i^{(l)} \| e_{iC}])
\end{align}
where $e_{iC} = \text{RBF}(\|\mathbf{r}_i - \mathbf{r}_C\|)$ encodes distances between atoms and cluster centers.

\subsection{PaiNN}
PaiNN maintains scalar ($h$) and vector ($\vec{v}$) features for equivariance. MCGM operates only on scalar features to preserve rotational equivariance:
\begin{align}
h^{(l+1)}, \vec{v}^{(l+1)} &= \text{PaiNN}(h^{(l)}, \vec{v}^{(l)}) \\
h^{(l+1)} &= h^{(l+1)} + W_{\text{dis}} \cdot [h_C^{(l)} \| e_{Ci}]
\end{align}
Vector features remain within their resolution level, while scalar features carry long-range information.

\subsection{DimeNet++}
DimeNet++ uses directional message passing with angular information. MCGM integrates at the node level after message aggregation:
\begin{align}
h^{(l+1)} &= h^{(l)} + \Delta h_{\text{DimeNet}}^{(l)} + W_{\text{dis}} \cdot [h_C^{(l)} \| e_{Ci}] \\
h_C^{(l+1)} &= h_C^{(l)} + W_{\text{agg}} \cdot \text{mean}_{i \in C}([h_i^{(l)} \| e_{iC}])
\end{align}
where $\Delta h_{\text{DimeNet}}^{(l)}$ includes the directional message passing with angular features. Edge features in DimeNet++ are updated internally within the model and do not directly interact with MCGM.

\subsection{GemNet-T and ViSNet}
Both GemNet-T and ViSNet employ sophisticated geometric features that require special handling to preserve their properties.

\textbf{GemNet-T} uses triplet-based message passing with angular information. MCGM operates on node embeddings while preserving edge-level angular features:
\begin{align}
h^{(l+1)}, m^{(l+1)} &= \text{GemNet-T}(h^{(l)}, m^{(l)}) \\
h^{(l+1)} &= h^{(l+1)} + W_{\text{dis}} \cdot [h_C^{(l)} \| e_{Ci}]
\end{align}

\begin{table}[htbp]
\centering
\caption{Key hyperparameters for MCGM experiments. Distance unit: 1 \AA{} = 0.1 nm.}
\label{tab:B1}
\begin{tabular}{lcc}
\toprule
\textbf{Parameter} & \textbf{OE62} & \textbf{AQM} \\
\midrule
Clustering iterations & 10 & 10 \\
Reduction ratio & 2 & 2 \\
Atom cutoff (\AA) & 6.0 & 4.0 \\
Cluster cutoff (\AA) & 4.0 & 4.0 \\
Learning rate & 1e-4 to 5e-4 & 1e-4 \\
Batch size & 8-64 & 32 \\
Early stopping patience & 50 & 150 \\
Loss function & L1 & MSE ($\lambda_E$=0.01, $\lambda_F$=0.99) \\
\bottomrule
\end{tabular}
\end{table}

\textbf{ViSNet} combines scalar and vector features with directional units. Similar to PaiNN, MCGM acts only on scalar features to maintain equivariance:
\begin{align}
h^{(l+1)}, \vec{v}^{(l+1)}, f^{(l+1)} &= \text{ViSNet}(h^{(l)}, \vec{v}^{(l)}, f^{(l)}) \\
h^{(l+1)} &= h^{(l+1)} + W_{\text{dis}} \cdot [h_C^{(l)} \| e_{Ci}]
\end{align}
where $f^{(l)}$ represents edge features with angular information.

\subsection{Key Hyperparameters}
The key hyperparameters used in our experiments are summarized in Table~\ref{tab:B1}.

\bibliographystyle{elsarticle-num}
\bibliography{refs.bib}
\end{document}